# Fault Detection of Broken Rotor Bar in LS-PMSM Using Random Forests

Juan C. Quiroz[a], Norman Mariun[b,c], Mohammad Rezazadeh Mehrjou[b,c], Mahdi Izadi[d], Norhisam Misron[c], Mohd Amran Mohd Radzi[b,c]

[a] Sunway University, Bandar Sunway, Selangor, Malaysia
[b] Centre of Advanced Power and Energy Research (CAPER), Universiti Putra Malaysia, 43400, Serdang, Selangor, Malaysia
[c] Department of Electrical and Electronic Engineering, Universiti Putra Malaysia, 43400, Serdang, Selangor, Malaysia
[d] Centre for Electromagnetic and Lightning Protection Research (CELP), Universiti Putra Malaysia, 43400, Serdang, Selangor, Malaysia

Corresponding author: Mohammad Rezazadeh Mehrjou,
Department of Electrical and Electronic Engineering, Universiti Putra Malaysia, 43400, Serdang, Selangor, Malaysia
Email address: mehrjou.mo@gmail.com

**Abstract**- This paper proposes a new approach to diagnose broken rotor bar failure in a line start-permanent magnet synchronous motor (LS-PMSM) using random forests. The transient current signal during the motor startup was acquired from a healthy motor and a faulty motor with a broken rotor bar fault. We extracted 13 statistical time domain features from the startup transient current signal, and used these features to train and test a random forest to determine whether the motor was operating under normal or faulty conditions. For feature selection, we used the feature importances from the random forest to reduce the number of features to two features. The results showed that the random forest classifies the motor condition as healthy or faulty with an accuracy of 98.8% using all features and with an accuracy of 98.4% by using only the mean-index and impulsion features. The performance of the random forest was compared with a decision tree, Naïve Bayes classifier, logistic regression, linear ridge, and a support vector machine, with the random forest consistently having a higher accuracy than the other algorithms. The proposed approach can be used in industry for online monitoring and fault diagnostic of LS-PMSM motors and the results can be helpful for the establishment of preventive maintenance plans in factories.

## 1. Introduction

Electrical motors convert electricity to mechanical energy. They account for two thirds of the total electricity use in industrial sites [1]. As a consequence, electrical machine manufacturers continuously strive to reduce the amount of energy used by motors. The standard IEC/EN 60034-30:2008 proposes IE4 as the highest efficiency for motors [2]. A LS-PMSM consists of a stator and a hybrid rotor. The rotor is comprised of an electricity conducting squirrel-cage and pairs of permanent magnet poles. The efficiency of LS-PMSMs stems from the combination of elements from permanent magnet synchronous motors and induction motors. The LS-PMSM provides (1) high efficiency, similar to permanent magnet synchronous motors, and (2) high starting torque, similar to induction motors [3].

Failures in electrical motors are common and difficult to prevent because motors are generally operated in industrial sites with different types of stress causing failures in various motor parts [4]. This has led to research on methods for early detection of failure in motors, to prevent motor inefficiencies and motor shutdown. In particular, rotor faults are significant because they exacerbate failures in other parts of the motor [5]. Various sensing techniques have been developed for broken rotor bar detection in electrical motors [5]. For instance, motor current signature analysis (MCSA) is a widely used technique due to its low cost and non-invasive nature [6]. In MCSA, the steady state current of a running motor is collected and recorded. From the recorded signal, features are extracted from the time domain, frequency domain, or time-frequency domain. These features are then used to make a diagnosis of the motor.

Fault analysis in induction motors has been widely applied. MCSA has been used to analyze faults in induction motors, such as rotor faults, bearing faults, eccentricity, misalignment, and stator faults [7–11]. Similar techniques have also been used to analyze vibration [12–16] and acoustic [17] signals of induction motors. The limitation of prior work is that most fault analysis has been applied to induction motors, electrical motors, fans, and gear boxes [7–17]. Yet, fault analysis in LS-PMSMs has been limited to a smaller set of faults, such as rotor faults, static eccentricity faults, and demagnetization [18–21]. Fault analysis in LS-PMSMs also suffers from a number of shortcomings: (1) the use of mathematical and simulated models to analyze faults, as opposed to using an LS-PMSM machine to collect data for fault detection; (2) the use of steady-state current for fault analysis; and (3) lack of machine learning algorithms for fault detection.

This paper makes three contributions. First, we used an LS-PMSM machine to collect current data while subjecting the motor to different loads. The rotor faults in our LS-PMSM machine were created by physically damaging the rotors of the LS-PMSM. We also analyzed the LS-PMSM starting with an initial load, as opposed to introducing a load after the motor had reached steady state, which is the common practice in prior work [7–11].

Second, we analyzed the transient current from when the motor is started. That is, we started the motor from standstill and waited for the motor to reach steady state, with the current from this transient period used for our analysis. Prior research uses the current from the steady state for fault analysis [7–11]. Finally, our third contribution is that our work is the first to apply machine learning for rotor fault detection in LS-PMSMs. We used random forests for the detection of rotor faults, and assessed the effectiveness of random forests by comparing with a decision tree, a Naïve Bayes classifier, logistic regression, linear ridge, and a support vector machine. To train these machine learning algorithms, we extracted thirteen time domain features from the transient current signal of the LS-PMSM, with the selection of the features based on prior work [22,23]. While machine learning methods have been used for fault detection in induction motors [7–11], to the best of our knowledge this is the first work to present fault analysis in LS-PMSMs by comparing various machine learning algorithms and using features extracted from the transient current signal to train and test these algorithms.

## 2. Fault Detection with Machine Learning

A random forest is a machine learning algorithm consisting of a number of independent decision trees [24]. A decision tree classifies an instance by testing attributes of the instance at each node of the tree [25]. Each node tests a particular attribute, with the leaves of the tree representing the output labels. Moving down a particular branch of a tree tests particular attributes at each node in order to arrive to an output label. A decision tree is typically built following a greedy approach, with the attribute/feature that results in the best split of the training data being used for the root node, and subsequently the attributes/features that result in the next best splits being used in the children nodes.

In contrast to a decision tree, a random forest uses bagging to build the decision trees in the forest [24]. In bagging, $T$ bootstrap sets are made by sampling with replacement $N$ training examples from the training set, with $T$ indicating the number of trees in the forest. Only 2/3 of each bootstrap set are used to build each tree, with the remaining 1/3, referred to as the out-of-bag data, used to get an estimate of the classification error of each tree. Fig. 1 illustrates the process of building a random forest with T trees.

In a typical decision tree, the greedy approach to building the tree can result in cases where the weaker features are not used at all. A random forest addresses this by choosing the best split in each node from a random subset of all the available features [24]. The random feature subset used for determining best node splits allows the weaker features to be represented in the random forest. Trees are grown to maximum length and without pruning to get low bias. Low correlation between the trees in the random forest is achieved by randomization as a result of the bootstrap samples and the random selection of features at each split [26]. Random forests have performed well in applications of fault diagnosis in rotating machinery [10,24].

Once the random forest has been built, an instance (x) is classified by passing the instance to each decision tree in the random forest (Fig. 2). Each decision tree classifies the instance by following a particular branch of the tree depending on the outcome from each node. The output of the random forest is then decided by taking the majority of the outputs from each tree. That is, the output of each tree is considered a vote, with the majority vote determining the output of the random forest.

Random forests provide a way to perform feature selection by using the importance of each feature derived while building the decision trees [24]. Every non-leaf node in a decision tree is a decision node which tests a single attribute, and based on the decision splits the data. Averaging the impurity decrease for each feature over all the trees in the random forest results in a score, this yields the importance for each feature. We used feature importances to reduce the number of features from 13 features to two features, as further described below.

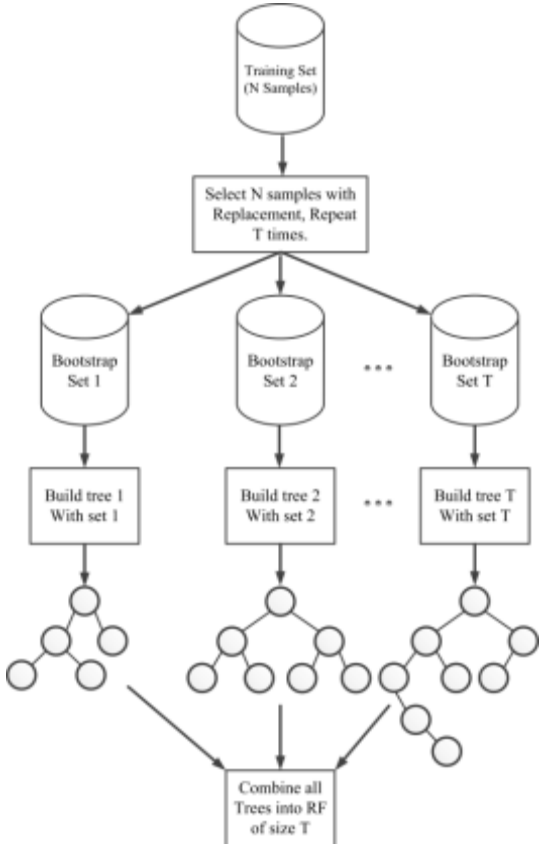

**Fig. 1.** Building a random forest with bagging

Random forests have been used for fault monitoring and diagnosis of induction motors and gear boxes, but not for LS-PMSMs. Niu et al. collected data from 21 sensors to detect vibration, current, voltage, and flux signals from an induction motor [7,8]. They extracted features from the time and frequency domains, and compared the performance of support vector machines, linear discriminant analysis, k nearest neighbors, random forests, and the adaptive resonance theory-Kohonen neural network, with random forests performing the best. In [9,10], random forests was used for fault detection, with the number of trees and the number of features selected at each node split optimized by a genetic algorithm. Karabadji et al. used the Weka machine learning library to compare random forests with various types of decision trees on the vibration signals of an industrial fan connected to an electrical motor [12,13]. They also used a genetic algorithm to optimize the type of tree to use and the choice of training and validation sets. In [17], Pandya et al. used acoustic signals for rolling element fault detection. They extracted features using Empirical Mode Decomposition, with modified k nearest neighbors outperforming random forests. Seera et al. proposed a hybrid model combining a Fuzzy Min-Max neural network (FMM) and a random forest [11]. They compared their hybrid model with FMM, a CART decision tree, and a hybrid FMM-CART ensemble. Cerrada et al. used random forests for multi-class fault diagnosis in spur gearboxes, with a genetic algorithm used to select the best subset of features out of 359 features in order to maximize diagnosis accuracy [14].

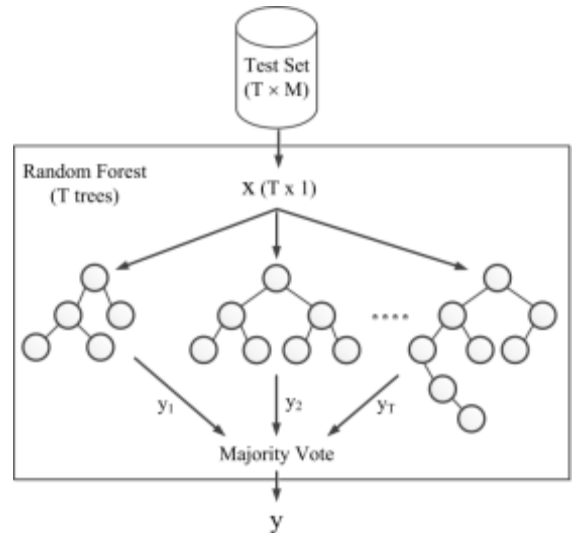

**Fig. 2.** Classifying an instance from the test set by passing the instance to each tree in the forest, and combining the outputs from all the trees using majority vote

Our work is the first to use the transient start up current for fault detection. This is in contrast to prior work which has used the steady state current for fault analysis [7–11]. In addition, all applications of random forests to fault detection have been limited to induction motors and they have not been applied to LS-PMSMs. The type of electrical motor has a significant influence on fault detection of the motor due to the differences in structure [18]. Thus, the fault monitoring from induction motors cannot be generalized to LS-PMSMs due to the differences in these two types of motors.

### 3. Line Start Permanent Magnet Synchronous Motors

An important obstacle for ordinary PMSMs is that they need an inverter for starting, which is not economical for many single speed applications. The squirrel-cage equipped permanent magnet motor-the LS-PMSM-provides a high efficiency motor with high starting capability, but without the need of a drive system [3,27]. LS-PMSMs can now reach super premium efficiency levels [28]. The structure of LS-PMSM comprises (1) a single or three-phase stator similar to an induction motor and (2) a hybrid rotor containing conductor bars and permanent magnets. The squirrel-cage bars in an electrical machine produce adequate high starting torque when the motor is run from standstill. Similar to asynchronous motors, squirrel-cage bars in LS-PMSM develop the startup performance during motor run up by enabling the rotor to have direct-on-line movement. When the load on the motor is unbalanced or the rotation speed is fluctuated, the squirrel-cage bars lessen the counter-rotating fields of the air gap, which otherwise would lead to significant losses [29].

Fig. 3 depicts the cross-section of one pole of a three-phase, four pole LS-PMSM. The rotor bar produces the starting torque as a result of the induction current in the bars establishing an electromagnetic field, which interacts with the rotatory field and subsequently pulls the motor toward the synchronism. Despite the induced current, the copper loss in the bars is negligible because of the synchronous operation.

Broken rotor bar is a major fault in squirrel cage motors. The presence of any damage in the rotor bar brings about secondary failures in the motor, leading to serious malfunctions of the motor. Once a rotor bar breaks, the condition of the neighboring bars also deteriorates over time due to the increased stress. Moreover, any defect in the bar influences the flux distribution. Thus, our work focuses on the detection of broken rotor bar faults.

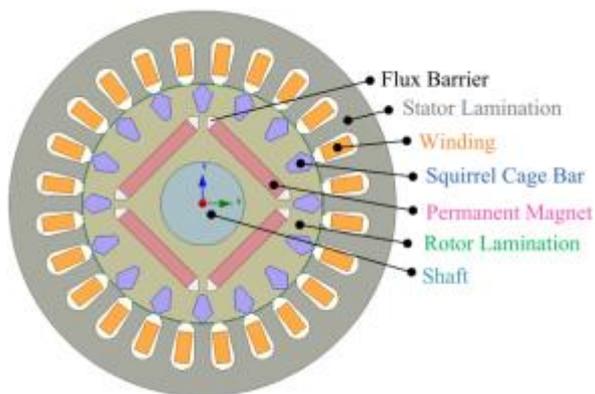

**Fig. 3.** The structure of three phase LS-PMSM (Four Pole).

Our work is most closely related to the work in [29,30]. In [30], Merjou et al. present broken rotor bar fault detection in LS-PMSMs, but their results are obtained from modeling with a simulation software using FEM at healthy condition and under fault. Merjou et al. do use transient current signals for their analysis, but these transient current signals are part of the simulation and not obtained from an actual LS-PMSM machine. Their analysis is limited to analyzing the current spectrum in the time domain by extracting statistical features. In [31], the Hilbert transform was used to extract the envelope of the current signal. From the envelope, time domain features were extracted, such as mean-index, RMS, skewness, kurtosis, among others. We improve on the work by Merjou et al. by conducting rotor bar fault detection on an LS-PMSM machine, whereas in [30-31] a software simulation is used for the fault analysis. We also take the additional step of including a machine learning algorithm trained with features from a healthy motor and a motor having a broken rotor bar. No machine learning method is presented in [30-31].

The remaining prior fault detection in LS-PMSMs includes eccentricity faults and demagnetization. In [18], Karami et al. analyze the eccentricity fault during the steady state operation of an LS-PMSM motor. The analysis is done using a software simulation of a three-phase LS-PMSM. Their work also uses analysis on the frequency domain using power spectral density (PSD) analysis. In [27], the irreversible demagnetization of an LS-PMSM is analyzed using transient analysis and the two-dimensional finite-element method. The analysis and results are based on a model, and not a physical LS-PMSM. In [21], the two-dimensional time-stepping finite element method is used to analyze the transient performance of a LS-PMSM simulated model. In [32], Lu et al. examined the behavior of an LS-PMSM during demagnetization condition along with the causes leading to the demagnetization. While a LS-PMSM machine is used to gather parameters and to understand the performance of the machine, the actual analysis is done on a mathematical model of the machine. Lu et al. were also one of the first to introduce machine learning, specifically a neural network, to detect demagnetization in LS-PMSM [32]. In [33–35], demagnetization in LS-PMSMs is also analyzed by using a simulation model.

In contrast to prior work, we used an LS-PMSM machine to collect current data for fault analysis, whereas prior work has solely used software simulation of LS-PMSMs. We also used transient current signal for our fault detection, as opposed to using the steady-state current. Lastly, prior fault analysis has studied demagnetization, whereas we focus on rotor faults.

## 4. Fault detection algorithm

Fig. 4 presents a flowchart of the fault detection method applied in this research. First,

the test workbench was set with the LS-PMSM having two different conditions: (1) healthy and (2) faulty—one broken rotor bar. The stator current signal was collected during the motor startup using a Hall-Effect current sensor. Next, the acquired current signals were preprocessed by truncating the data to the first 40 periods of the current signal. The data was also cleaned by ensuring that the start and stop points of the data fell in zero. In the next stage, 13 time domain features were extracted (discussed below) from the current signal. Finally, the features were used to train and test the random forest.

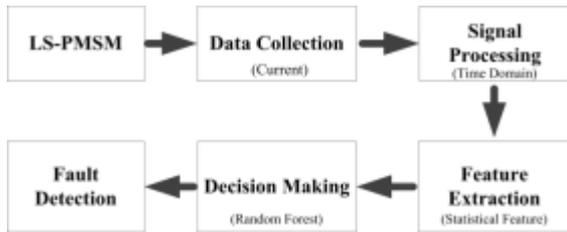

**Fig. 4.** The flowchart diagram of fault diagnosis system.

### 4.1 Setup Configuration

Fig. 5 shows the schematic of the test facilities used to examine the broken rotor bar fault. The system consisted of a three phase power supply, an LS-PMSM, a torque and speed sensor, mechanical couplings, a magnetic powder brake acting as a load, a DC source, a Hall-Effect current sensor, an oscilloscope, and a computer.

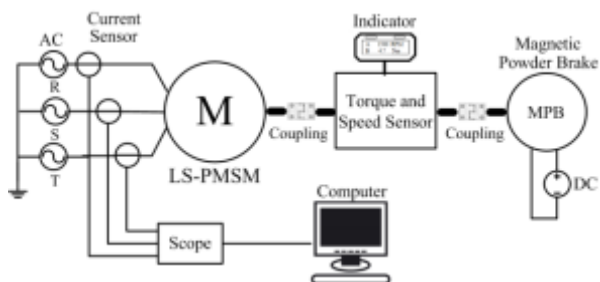

**Fig. 5.** Schematic of the LS-PMSM test facility.

Experimental tests were performed on a two-pole-pair LS-PMSM to evaluate the accuracy of fault detection achieved through the integration of time domain features and a random forest algorithm. Experimental data was directly obtained from the test motors by using the test bench shown in Fig. 6(a). The magnetic powder brake was coupled to the LS-PMSM to produce four different levels of starting loads: 0 Nm, 0.5 Nm, 1.0 Nm, and 1.5 Nm. The brake was controlled with current from the DC power supply. The broken rotor bar fault was made by drilling into the rotor bars, with an example shown in Fig. 6(b). Table 2 lists the parameters of the LS-PMSM used in this research.

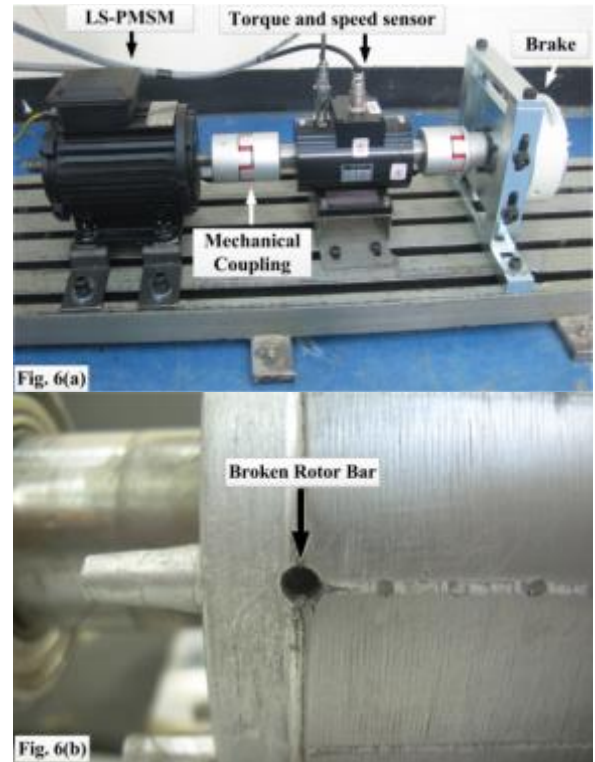

**Fig. 6.** (a) Experimental test bench (b) Rotor with one bar breakage.

**Table 2**
Motor Specifications.

| Parameters | Value |
| --- | --- |
| Rated Output Power (HP) | 1 |
| Rated Voltage (V) | 415 |
| Rated Frequency (Hz) | 50 |
| Rated Torque (N.m) | 4.8 |
| Rated Speed (rpm) | 1500 |
| Rated Current (A) | 1.3 |
| Starting Torque | 2.3 |
| Number of Poles | 4 |
| Connection | Y |
| Number of Stator Slots | 24 |
| Number of Rotor Slots | 16 |

In fault monitoring of motors, stator current signal is commonly used due to its noninvasive nature, making it easy to measure and analyze in industry. In LS-PMSM, the squirrel-cage bars carry appreciable current during the startup stage, while little or no current when the machine operates in steady-state [36]. Thomson et al. proposed monitoring and analyzing the stator current during motor startup [37]. Transient current is an accurate method for detection of the failure in an induction machine [37,38]. Broken bars generate extra components in the stator current that depend on the rotor speed, and these can be measured during the motor start up. However, these extra components fade away or coincide with other components, like saturation-induced components that do not contain information related to the fault, when the motor reaches steady state [39]. One of the challenges of using the current during startup, and one of the main reasons why it has not been commonly used for fault detection, is the complexity of extracting fault characteristics as the startup current is extremely unsteady and its period is quite short [40]. Recent analysis of transient current signature in induction motors [28,29] has shown that transient stator current is independent of load conditions that make it suitable for fault detection. In [30,31], the effect of load for fault detection in induction motors during the transient state was not considered. However, in LS-PMSMs the starting torque decreases whenever there is a broken rotor bars [32]. In this paper, we also consider the effect of load on the starting time.

The current signal data was acquired with the Pico Scope 4424 oscilloscope and its accompanying software. For each test, the current signal in one phase was recorded. The stator current signal is suitable for the acquisition system, which employs a Hall sensor (LEM-LTS25-NP) for this signal condition. A sampling rate of 5 kHz and a resolution of 12b were used for recording the signals. The number of 40 periods of startup current signal was selected for feature calculation, with approximately 4K sampled data used. We used the period of signal instead of sampled data because the fundamental frequency is not fixed in the normal electrical supply. To analyse the data, both the transient state and a portion of the steady state of the current signal were considered.

Forty tests were performed for each condition. A total of 320 tests were done based on healthy and faulty motor with four different levels of starting load. The torque was measured by a Dacell-TRB-10Ktorque meter, with the speed measured by a MP-981. Fig. 7 shows the current signal extracted from healthy and faulty motors running under high load conditions.

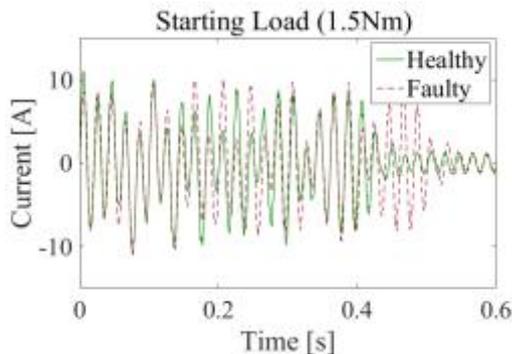

**Fig. 7.** Experimental current signal in high-load condition for a healthy and a faulty motor.

*4.2 Signal Processing and Feature Calculation*

After preprocessing, those features related to broken rotor bar was extracted from the raw signal (time domain signal). Statistical features like average, variance, skewness, and kurtosis were used as a quick test for changes in the pattern of signals. These parameters are commonly used for statistical analysis of the current and vibration signals in the time domain [41]. If variation in the condition of the motor causes a change in the current signal, monitoring this signal can provide detection information. The statistical features may be dimensional or non-dimensional. Dimensional parameters include mean-index, Root Mean Square (RMS), Root-sum-of-squares level (RSS), Peak–Peak value and Energy. Non-dimensional parameters include pulse index, waveform index (Shape Factor), impulsion index, peak index (Crest factor), tolerance index (Margin factor), skewness index, and kurtosis index [42]. Han and et al. compared the features obtained from time domain analysis of the steady state current signal and indicated the ability and efficiency of these features for detection of different faults [43]. The definitions of the features used in current work are the following:

**Dimensional parameters**
Mean-index
$$X_{Mean} = \frac{1}{N}\sum_{i=1}^{N} X_i \quad (1)$$
RMS
$$X_{RMS} = \sqrt{\frac{1}{N}\sum_{i=1}^{N} X_i^2} \quad (2)$$
RSSQ
$$X_{RSS} = \sqrt{\sum_{i=1}^{N} |X_i^2|} \quad (3)$$
Peak to peak
$$X_{PP} = \max(X) - \min(X) \quad (4)$$
Energy
$$X_{Energy} = \sum_{i=1}^{N} X_i^2 \quad (5)$$

**Non-dimensional parameters:**
Shape Factor
$$X_{Wi} = \frac{X_{RMS}}{\frac{1}{N}\sum_{i=1}^{N}|X_i|} \quad (6)$$
Impulsion
$$X_{Ii} = \frac{\max|X|}{\frac{1}{N}\sum_{i=1}^{N}|X_i|} \quad (7)$$
Crest factor
$$X_{Pi} = \frac{\max|X|}{X_{RMS}} \quad (8)$$
Margin factor
$$X_{Ti} = \frac{\max|X|}{(\frac{1}{N}\sum_{i=1}^{N}|X_i|^{1/2})^2} \quad (9)$$

Peak-to-average power ratio
$$X_{PA} = (max(X))^2 / (X_{RMS})^2 \quad (10)$$
Variance
$$X_{Va} = \frac{1}{N-1}\sum_{i=1}^{N}|X_i - X_{Mean}|^2 \quad (11)$$
Skewness
$$X_{Si} = \frac{1}{N}\sum_{i=1}^{N}\frac{(X-X_{Mean})^3}{\sigma^3} \quad (12)$$
Kurtosis
$$X_{Ki} = \frac{1}{N}\sum_{i=1}^{N}\frac{(X-X_{Mean})^4}{\sigma^4} \quad (13)$$

For the equations presented, X is a signal, N is number of sampled data points of the signal, and σ is the standard deviation calculated as follows:

$$\sigma = \sqrt{\frac{1}{N}\sum_{i=1}^{N}(X-X_{Mean})^2} \quad (14)$$

*4.3 Experimental Setup for Classification*

We used the Python Scikit-learn library for the random forest implementation [44], to conduct the test and training, and for the comparison with the other machine learning algorithms. In this implementation, the random forest combines the decision trees by averaging their probabilistic prediction, instead of having each decision tree vote for a single class [44]. We used the Gini impurity to measure the quality of a split, and selected among $\sqrt{p}$ maximum random features for each node split, where p is the total number of features. Random forests were tested with 10, 100, 200, 500, and 1000 trees. We further compared the performance of the random forest to a single CART decision tree [25], Gaussian Naïve Bayes classifier [45], logistic regression [46], linear ridge classifier [47], and a support vector machine (SVM) with a radial basis function (RBF) kernel [48]. For all experiments, we measured out of sample performance using the Area under the ROC Curve (AUC) from 5-fold cross-validation. After training the random forest, we used the feature importances for feature selection [14,49]. The results show how the accuracy of the random forest changed as a factor of the number of features used, starting with the 13 maximum features and subsequently removing the less important features.

**5. Results and Discussion**

Table 3 presents the results of a random forest algorithm trained with different feature sets and number of trees in the forest. First, we used all 13 features to train and test the random forests. The performance for a random forest with 100 trees was 99.6%. However, the performance remains consistent independent of the number of decision trees. Next, we reduced the number of features by using the feature importances of the random forest. Fig. 8 shows a plot of the importance of each of the 13 features. The three most informative features were mean-index, impulse factor, and shape factor. These three features were used to train and test another set of random forests. The performance of a random forest with 100 trees and using these three features was also 99.6%, the same as when using all 13 features. In addition, the performance of the random forest when using three features is independent of the number of trees in the forest. Lastly, using only the two top features, mean-index and impulse factor, gives results comparable to using only three features.

**Table 3.**
The accuracy results of random forest for broken rotor bar detection based on different number of features. Means and standard deviations (STD) over folds are reported.

| | Features | All features | Mean-index, Impulsion, Shape Factor | Mean-index, Impulsion |
|---|---|---|---|---|
| Accuracy | 10 Trees mean | 99.5 | 99.3 | 99.3 |
| | STD | 0.007 | 0.008 | 0.012 |
| | 100 Trees mean | 99.6 | 99.6 | 99.6 |
| | STD | 0.007 | 0.006 | 0.006 |
| | 200 Trees mean | 99.8 | 99.6 | 99.6 |
| | STD | 0.003 | 0.006 | 0.006 |
| | 500 Trees mean | 99.8 | 99.6 | 99.6 |
| | STD | 0.003 | 0.006 | 0.006 |
| | 1000 Trees mean | 99.7 | 99.6 | 99.6 |
| | STD | 0.005 | 0.006 | 0.006 |

Given that two features are sufficient to yield high performances and that the performance of the random forest is independent of the number of trees, we hypothesized whether all the trees in the random forest were the same when using only two features. To test this hypothesis, we trained and tested a single CART decision tree using different random seeds, and examined the resulting topologies of the decision trees.

Figure 9 illustrates two example decision trees generated using different random seeds. The first decision tree (Figure 9(a)) has depth four and a total of nine nodes. The decision nodes of the first decision tree test different values of the mean-index and the impulse factor to determine whether the motor is healthy or faulty. For example, following one path of the tree indicates that if the mean-index of the startup current sample is less than or equal to -0.269, then the motor is faulty. Another path of the decision tree shows that if the mean is greater than -0.269, the impulse factor is less than or equal to 0.4642, and the mean-index is less than 0.3446, then the motor is healthy. In contrast, the second decision tree (Figure 9(b)) has depth two and a total of five nodes. In particular, the second decision tree relies solely on the mean-index feature to determine whether the motor is faulty or healthy. This shows that even when using only two features, the random forest will contain diverse trees which make it less likely to over fit to the training set.

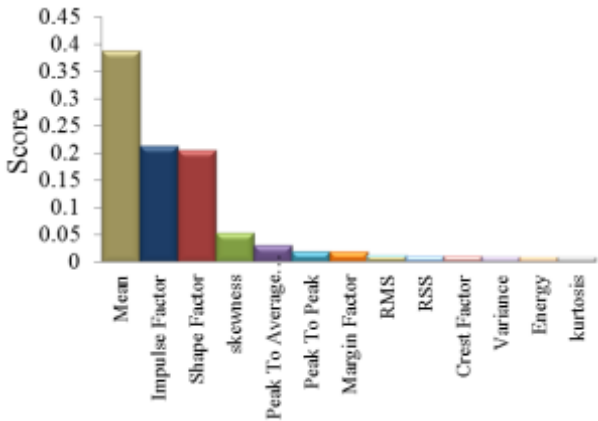

**Fig. 8.** Feature importance for the 13 features.

To assess the effectiveness of random forest, we tested baseline algorithms on our transient current data set from the LS-PMSM. The AUC for the random forest (with 100 trees) compared to the other algorithms is reported in Table 4. We tested all the algorithms using all 13 features, using only three features (mean-index, impulsion factor, and shape factor), and using only two features (mean-index and impulsion factor). The random forest outperforms all of the other classifiers. For some of the methods, such as CART and Naïve Bayes, reducing the number of features leads to minimal drop in performance. For the SVM, reducing the number features results in a minimal increase in performance.

The approach presented here can detect whether the LS-PMSM is healthy or faulty by using a random forest to analyze transient current signal data from the LS-PMSM. Current signature analysis has the advantage of being non-invasive. In addition, feature importances derived from building the random forest was be used to reduce the number of features, with the random forest still maintaining high prediction accuracy. The performance of the random forest was independent of the number of trees in the forest.

Finally, even though the random forest performed the best, all the other methods performed well with accuracies of over 90%.

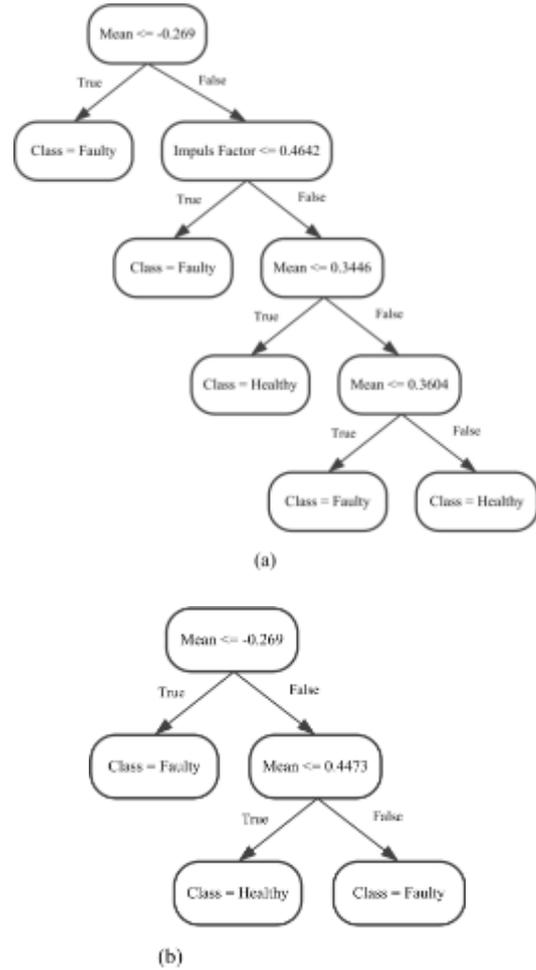

**Fig. 9.** (a) CART decision tree built and trained using mean-index and impulse factor, and (b) CART decision tree built and trained using mean-index and impulse factor, but only mean-index is used in the nodes of the decision tree.

**Table 4.**
AUC comparison for broken rotor bar data, means and standard deviations (STD) over folds are reported.

| Classifier | All features | | Mean-index, Impulsion, Shape Factor | | Mean-index, Impulsion | |
|---|---|---|---|---|---|---|
| | mean | STD | mean | STD | mean | STD |
| Random Forest | **99.6** | **0.01** | **99.6** | **0.01** | **99.6** | **0.01** |
| CART | 98.8 | 0.01 | 97.5 | 0.03 | 96.9 | 0.04 |
| Naïve Bayes | 96.9 | 0.05 | 96.3 | 0.07 | 95 | 0.1 |
| Logistic regression | 94 | 0.09 | 93.8 | 0.1 | 93.8 | 0.1 |
| Linear Ridge | 93.9 | 0.1 | 93.8 | 0.1 | 93.8 | 0.1 |
| SVM | 93.8 | 0.1 | 94.4 | 0.1 | 95.2 | 0.1 |

In this case study, random forest outperformed the other classifiers. A random forest combines the predictions of the decision trees in the forest by averaging their predictions. This combination of the individual estimations made by the trees in the forest lowers the variance of the model, which gives it a predictive advantage over a single, simple classifier [24].

## 6. Conclusions

In this paper, we presented a new approach for the fault detection of rotor bars in LS-PMSM using random forests. We collected the transient current signal during the startup phase of a healthy LS-PMSM machine and an LS-PMSM machine with rotor faults. During the data collection, the LS-PMSM machines were subjected to different load conditions. Experimental results indicate the validity and reliability of the random forest fault detection method. The approach attains a high correct rate of diagnosis of 98.8%. The accuracy of the random forest was found to be independent of the number of trees. With feature importances, we identified mean-index and impulsion as the two most important features. With these two features alone, the random forest achieved a performance of 98.4%. Comparing the random forest to other machine learning algorithms showed that random forest performed the best. However, all algorithms achieved accuracies over 90%, which gives practitioners the confidence and flexibility to choose various types of algorithms and not be limited to random forests.

**Acknowledgments**

The authors would like to express their gratitude to Ministry of Higher Education Malaysia for financial support through grant number FRGS-5524356 and Universiti Putra Malaysia for the facilities provided during this research work.